\documentclass[twoside,11pt]{article}

\usepackage{blindtext}
\usepackage{tabularx}
\usepackage[table]{xcolor}
\usepackage{url}
\usepackage{jmlr2e}
\usepackage{graphicx}
\usepackage{booktabs}
\usepackage{multirow}
\usepackage{amsmath}
\usepackage{lastpage}

\jmlrheading{23}{2022}{1-\pageref{LastPage}}{1/21; Revised 5/22}{9/22}{21-0000}{Professor Cline}

\ShortHeadings{pyFAST: A Modular PyTorch Framework for Time Series Modeling}{pyFAST: A Modular PyTorch Framework for Time Series Modeling}
\firstpageno{1}

\usepackage{enumerate}

\begin{document}

\title{pyFAST: A Modular PyTorch Framework for Time Series Modeling with Multi-source and Sparse Data}

\author{
	\name Zhijin Wang \email zhijin@jmu.edu.cn \\
	\addr College of Computer Engineering, Jimei University, 361021 Xiamen, China
	\AND
        \name Senzhen Wu \email szwbyte@gmail.com \\
	\addr College of Computer Engineering, Jimei University, 361021 Xiamen, China
	\AND
        \name Yue Hu \email yuehu@jmu.edu.cn \\
        \addr Chengyi College, Jimei University, 361021 Xiamen, China
        \AND
	\name Xiufeng Liu \email xiuli@dtu.dk \\
	\addr Department of Technology, Management and Economics, Technical University of Denmark, 2800 Kgs. Lyngby, Denmark
}

\editor{}

\maketitle

\begin{abstract} 
Modern time series analysis demands frameworks that are flexible, efficient, and extensible. However, many existing Python libraries exhibit limitations in modularity and in their native support for irregular, multi-source, or sparse data. We introduce \texttt{pyFAST}, a research-oriented PyTorch framework that explicitly decouples data processing from model computation, fostering a cleaner separation of concerns and facilitating rapid experimentation. Its data engine is engineered for complex scenarios, supporting multi-source loading, protein sequence handling, efficient sequence- and patch-level padding, dynamic normalization, and mask-based modeling for both imputation and forecasting. \texttt{pyFAST} integrates LLM-inspired architectures for the alignment-free fusion of sparse data sources and offers native sparse metrics, specialized loss functions, and flexible exogenous data fusion. Training utilities include batch-based streaming aggregation for evaluation and device synergy to maximize computational efficiency. A comprehensive suite of classical and deep learning models (Linears, CNNs, RNNs, Transformers, and GNNs) is provided within a modular architecture that encourages extension. Released under the MIT license at \href{https://github.com/freepose/pyFAST}{GitHub}\footnote{\url{https://github.com/freepose/pyFAST}}, \texttt{pyFAST} provides a compact yet powerful platform for advancing time series research and applications.
\end{abstract}

\section{Introduction}
Time series analysis is a foundational task in domains ranging from financial forecasting and healthcare monitoring to climate science and industrial predictive maintenance. In response, a variety of Python libraries have been developed for forecasting, classification, and representation learning \citep{pro/acm2016/1Abadi, pro/nips2019/8024Paszke, art/jmlr2020/1Alexandrov, art/corr2019/Markus, art/jmlr2020/1Tavenard}. While these toolkits have significantly advanced the field, they often present challenges in modularity, native handling of sparse data, and support for heterogeneous or alignment-free multi-source integration. For instance, many frameworks tightly couple data pipelines with model implementations, which can hinder flexible experimentation. Others are optimized for specific tasks or model families, creating silos that impede systematic benchmarking across diverse architectural paradigms. These gaps are particularly salient in real-world applications where data are inherently irregular, originate from multiple asynchronous sources, or contain missing values—scenarios that demand not only reproducibility and extensibility but also a powerful capacity to fuse heterogeneous data and rapidly prototype novel models. The absence of a unified framework that addresses these challenges simultaneously forces researchers into cumbersome, ad-hoc workflows, slowing the pace of innovation.

To address these needs, we present \texttt{pyFAST}, a modular and efficient framework built on PyTorch. \texttt{pyFAST} is designed with a clear separation between data processing and model computation. It provides native support for sparse and multi-source time series, including LLM-inspired architectures for alignment-free data fusion, and incorporates a suite of efficient, tensor-based utilities for dynamic data handling. By combining these features with a comprehensive model library and an extensible benchmarking suite, \texttt{pyFAST} enables rapid experimentation on complex real-world challenges, establishing itself as a robust platform for modern time series research.

\begin{figure}[t!]
	\centering
	\includegraphics[width=0.95\textwidth]{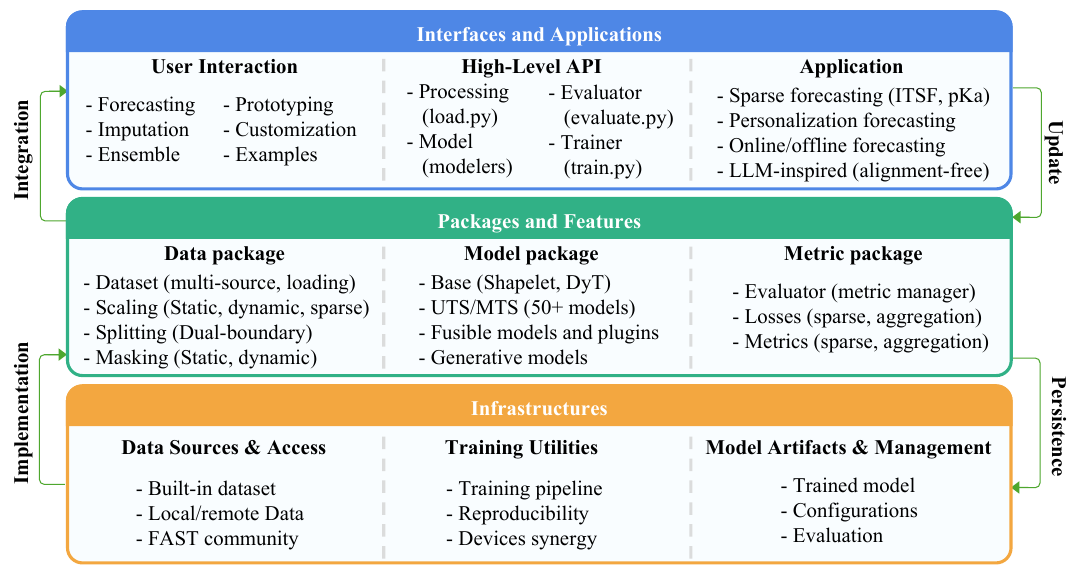}
	\vspace{-10pt}
	\caption{Architectural overview of the \texttt{pyFAST} library.\label{fig:software_overview}}
	\vspace{-15pt}
\end{figure}

\section{Software Overview}
As illustrated in Figure \ref{fig:software_overview}, \texttt{pyFAST} is architected in three distinct layers: Interfaces and Applications, Packages and Features, and Infrastructures. The framework's design adheres to a strict modular principle to ensure flexibility and extensibility.

A central innovation is the \texttt{Data} package, which provides advanced data handling capabilities. It supports loading from multiple sources, dynamic scaling, and integrating datasets without requiring temporal alignment. This alignment-free capability is achieved by tokenizing each data stream independently and using source-specific identifiers, allowing a downstream model to learn cross-modal relationships. The package also features efficient sequence- and patch-level padding, high-speed dynamic normalization during training, and versatile mask modeling strategies for imputation and forecasting. These tensor-based padding and patching operations are performed dynamically at the batch level, which significantly reduces memory overhead and preprocessing time compared to static, dataset-level approaches. For bioinformatics applications, it supports dynamic padding for variable-length sequences to enable high-throughput processing.

The \texttt{Model} package hosts a comprehensive suite of models, from classical statistical baselines to modern deep learning architectures including CNNs, RNNs, Transformers, and Graph Neural Networks. All models are implemented using optimized tensor operations to ensure scalability and speed. A \texttt{Base} submodule provides reusable building blocks, such as shapelet layers, time series decomposition methods, and DyT, which can be composed to construct novel architectures with minimal boilerplate code.

The \texttt{Training Utilities} package streamlines the experimental workflow. The \texttt{Trainer} class automates training loops with integrated support for checkpointing, early stopping, and learning rate scheduling. The \texttt{metric} module includes both standard and sparse metrics tailored to scenarios with missing data. Finally, the \texttt{Evaluator} class enables consistent benchmarking, featuring batch-based stream aggregation for efficient large-scale assessments. This streaming evaluator computes metrics on-the-fly for each batch, enabling robust evaluation on datasets that are too large to fit into main memory. To further promote reproducible research, the framework is accompanied by a collection of curated benchmark datasets.\footnote{The benchmark datasets are available at \url{https://zenodo.org/communities/fast}} This explicit decoupling of data processing from model computation enhances clarity and maintainability, allowing researchers to focus on novel modeling strategies while relying on a robust and reproducible data pipeline.

\section{Comparison with Related Work}

The time series ecosystem includes many specialized libraries. TensorFlow Time Series \citep{pro/acm2016/1Abadi} is tightly integrated with TensorFlow, while GluonTS \citep{art/jmlr2020/1Alexandrov} excels at probabilistic forecasting. PyTorch Forecasting \citep{pro/nips2019/8024Paszke} offers strong utilities for interpretability. Within the scikit-learn ecosystem, sktime \citep{art/corr2019/Markus} provides a unified interface for general-purpose time series analysis, and tslearn \citep{art/jmlr2020/1Tavenard} focuses on similarity-based methods. More recently, libraries like StatsForecast, NeuralForecast \citep{garza2022statsforecast, olivares2022library_neuralforecast}, and TSLib \citep{pro/iclr2023/1Wu} have emphasized highly efficient and scalable model implementations. Table \ref{tab:library_comparison} provides a comparative overview.

\begin{table*}[t!]
	\centering
	\vspace{-10pt}
	\caption{Comparison of \texttt{pyFAST} with representative time series libraries.}
	\label{tab:library_comparison}
	\rowcolors{2}{white}{gray!15}
	{\scriptsize
		\begin{tabularx}{\textwidth}{@{}p{1.7cm}p{3.1cm}p{2.1cm}p{2.1cm}p{2.1cm}p{2.1cm}@{}}
			\toprule
			\textbf{Feature} & \textbf{\texttt{pyFAST}} & \textbf{GluonTS} & \textbf{PyTorch Forecasting} & \textbf{sktime} & \textbf{TSLib} \\
			\midrule
			Modularity & \textbf{High} (Component-based, highly customizable; data–model decoupling) & Medium (Model zoo, some customization) & Medium (Layer-based, some customization) & High (Class-based, extensible framework) & Medium (Modular design) \\
			Extensibility & \textbf{High} (Easy to add custom models/components; flexible pipelines) & Medium (Adding new models requires effort) & Medium (Custom layers supported) & High (Extensible class hierarchy, well-defined interfaces) & Medium (Extensible, but might require deeper code modification) \\
			Model Breadth (DL \& Classical) & \textbf{Extensive} (Deep learning \& classical statistical models) & Broad (Strong DL, limited classical) & Broad (Classical ML and deep learning) & Broad (Classical statistical \& ML models) & Broad (Focus on deep learning, but includes classical) \\
			Transformer Models & \textbf{Extensive} (PatchTST, Informer, Autoformer, diverse variants) & Limited (Basic Transformer encoder-decoder) & Yes (Time series Transformer layers) & Limited (No dedicated Transformer models) & Yes (Transformer models included) \\
			LLM-Inspired Models \& Multi-source Support & \textbf{Yes} (Transformer-based LLM adaptations, \textbf{native multi-source support without alignment}) & No & No & No & No \\
			Sparse Data Support & \textbf{Yes} (\textbf{native sparse data support, sparse metrics and losses}) & Limited & Limited & Limited & Limited \\
			Exogenous Data Fusion & \textbf{Yes} (\textbf{flexible exogenous data integration}) & Supported & Supported & Supported & Supported \\
			Sequence Learning Focus & \textbf{Yes} (\textbf{batch-wise dynamic padding, stream aggregation evaluation, dynamic segmentation/patching}) & No & No & No & No \\
			Imputation & \textbf{Yes} (Dedicated imputation models/methods) & Yes (Probabilistic models can be adapted) & Limited (Not a primary focus) & Yes (Comprehensive toolbox includes imputation) & Limited (Not explicitly highlighted) \\
			Benchmarking Suite & \textbf{Yes} (Standardized benchmarking framework; reproducible evaluation) & Limited (Basic benchmarks) & No & Yes (Comprehensive evaluation framework) & Yes (Benchmarking scripts provided) \\
			Focus & \textbf{Research \& Prototyping} (Flexibility, customization, advanced data–model decoupling) & \textbf{Probabilistic Forecasting} (Scalability and uncertainty estimation) & \textbf{Explainability} (Interpretability and time series-specific layers) & \textbf{General Time Series Analysis} (Broad toolbox for diverse tasks) & \textbf{Efficiency \& Reproducibility} (Emphasis on performance and code clarity) \\
			\bottomrule
		\end{tabularx}
	}
	\vspace{-10pt}
\end{table*}

Despite these advances, many libraries tightly couple data processing with model definitions, lack explicit designs for heterogeneous or sparse data, and provide limited facilities for advanced sequence modeling. While existing libraries excel at specific tasks like probabilistic forecasting or efficiency, \texttt{pyFAST} is uniquely positioned as a research-centric framework designed for the messy, heterogeneous data characteristic of many real-world scientific and industrial domains. Compared to TSLib, which prioritizes efficient implementations, \texttt{pyFAST} extends functionality with LLM-inspired architectures, dedicated sparse metrics, and flexible exogenous data fusion, making it particularly suited for advanced research applications. 

\section{Conclusion}
\texttt{pyFAST} delivers a modular framework that separates data processing from model computation, providing native support for sparse data, alignment-free multi-source integration, and LLM-inspired architectures. Its efficient dynamic padding, normalization, and mask modeling enable broad applicability across health/healthcare, energy load/consumption, and protein sequence modeling. Released under MIT license, \texttt{pyFAST} offers a high-performance platform promoting cutting-edge research and agile united process in time series analysis.

\bibliography{fast}

\begin{thebibliography}{8}
\providecommand{\natexlab}[1]{#1}
\providecommand{\url}[1]{\texttt{#1}}
\expandafter\ifx\csname urlstyle\endcsname\relax
  \providecommand{\doi}[1]{doi: #1}\else
  \providecommand{\doi}{doi: \begingroup \urlstyle{rm}\Url}\fi

\bibitem[Abadi(2016)]{pro/acm2016/1Abadi}
Mart{\'\i}n Abadi.
\newblock Tensorflow: learning functions at scale.
\newblock In \emph{Proceedings of the 21st International Conference on Functional Programming}, pages 1--1, Nara, Japan, September 2016. {ACM}.
\newblock \doi{10.1145/2951913.2976746}.

\bibitem[Alexandrov et~al.(2020)Alexandrov, Benidis, Bohlke-Schneider, Flunkert, Gasthaus, Januschowski, Maddix, Rangapuram, Salinas, Schulz, et~al.]{art/jmlr2020/1Alexandrov}
Alexander Alexandrov, Konstantinos Benidis, Michael Bohlke-Schneider, Valentin Flunkert, Jan Gasthaus, Tim Januschowski, Danielle~C Maddix, Syama Rangapuram, David Salinas, Jasper Schulz, et~al.
\newblock Gluonts: Probabilistic and neural time series modeling in python.
\newblock \emph{Journal of Machine Learning Research}, 21\penalty0 (116):\penalty0 1--6, June 2020.
\newblock \doi{10.48550/arXiv.1906.05264}.

\bibitem[Garza et~al.(2022)Garza, Canseco, Challú, and Olivares]{garza2022statsforecast}
Azul Garza, Max~Mergenthaler Canseco, Cristian Challú, and Kin~G. Olivares.
\newblock {StatsForecast}: Lightning fast forecasting with statistical and econometric models.
\newblock {PyCon} Salt Lake City, Utah, US 2022, 2022.
\newblock URL \url{https://github.com/Nixtla/statsforecast}.

\bibitem[L{\"o}ning et~al.(2019)L{\"o}ning, Bagnall, Ganesh, Kazakov, Lines, and Kir{\'a}ly]{art/corr2019/Markus}
Markus L{\"o}ning, Anthony Bagnall, Sajaysurya Ganesh, Viktor Kazakov, Jason Lines, and Franz~J Kir{\'a}ly.
\newblock sktime: A unified interface for machine learning with time series.
\newblock \emph{Computing Research Repository}, September 2019.
\newblock \doi{10.48550/arXiv.1909.07872}.

\bibitem[Olivares et~al.(2022)Olivares, Challú, Garza, Canseco, and Dubrawski]{olivares2022library_neuralforecast}
Kin~G. Olivares, Cristian Challú, Azul Garza, Max~Mergenthaler Canseco, and Artur Dubrawski.
\newblock {NeuralForecast}: User friendly state-of-the-art neural forecasting models.
\newblock {PyCon} Salt Lake City, Utah, US 2022, 2022.
\newblock URL \url{https://github.com/Nixtla/neuralforecast}.

\bibitem[Paszke et~al.(2019)Paszke, Gross, Massa, Lerer, Bradbury, Chanan, Killeen, Lin, Gimelshein, Antiga, et~al.]{pro/nips2019/8024Paszke}
Adam Paszke, Sam Gross, Francisco Massa, Adam Lerer, James Bradbury, Gregory Chanan, Trevor Killeen, Zeming Lin, Natalia Gimelshein, Luca Antiga, et~al.
\newblock Pytorch: An imperative style, high-performance deep learning library.
\newblock In \emph{Proceedings of the 33rd Advances in Neural Information Processing Systems}, volume~32, pages 8024--8035, Vancouver, BC, Canada, June 2019.
\newblock \doi{10.48550/arXiv.1906.05264}.

\bibitem[Tavenard et~al.(2020)Tavenard, Faouzi, Vandewiele, Divo, Androz, Holtz, Payne, Yurchak, Ru{\ss}wurm, Kolar, et~al.]{art/jmlr2020/1Tavenard}
Romain Tavenard, Johann Faouzi, Gilles Vandewiele, Felix Divo, Guillaume Androz, Chester Holtz, Marie Payne, Roman Yurchak, Marc Ru{\ss}wurm, Kushal Kolar, et~al.
\newblock Tslearn, a machine learning toolkit for time series data.
\newblock \emph{Journal of Machine Learning Research}, 21\penalty0 (118):\penalty0 1--6, January 2020.
\newblock \doi{10.5555/3455716.3455834}.

\bibitem[Wu et~al.(2023)Wu, Hu, Liu, Zhou, Wang, and Long]{pro/iclr2023/1Wu}
Haixu Wu, Tengge Hu, Yong Liu, Hang Zhou, Jianmin Wang, and Mingsheng Long.
\newblock Timesnet: Temporal 2d-variation modeling for general time series analysis.
\newblock In \emph{Proceedings of the 11th International Conference on Learning Representations}, pages 1--5, Kigali, Rwanda, May 2023. OpenReview.net.
\newblock \doi{10.48550/arXiv.2210.02186}.

\end{thebibliography}

\end{document}